\pdfoutput=1

\documentclass[11pt]{article}

\usepackage[final]{acl}

\usepackage{times}
\usepackage{latexsym}
\usepackage[most]{tcolorbox}
\tcbuselibrary{breakable}
\usepackage[T1]{fontenc}

\usepackage[utf8]{inputenc}
\usepackage[shortlabels]{enumitem}

\usepackage{microtype}
\usepackage{graphicx}
\usepackage{linguex}
\usepackage{subcaption}
\usepackage{amssymb}

\usepackage{multicol}
\usepackage{multirow} 
\usepackage{booktabs}

\RequirePackage{etoolbox}
\RequirePackage{xcolor}
\definecolor{Gray}{gray}{0.9}
\definecolor{cb-black}      {RGB}{ 0,   0,   0}
\definecolor{cb-blue-green} {RGB}{ 0,  073,  073}
\definecolor{cb-green-sea}  {RGB}{ 0, 146, 146}
\definecolor{cb-rose}       {RGB}{255, 109, 182}
\definecolor{cb-salmon-pink}{RGB}{255, 182, 119}
\definecolor{cb-purple}     {RGB}{ 73,   0, 146}
\definecolor{cb-blue}       {RGB}{ 0, 109, 219}
\definecolor{cb-lilac}      {RGB}{182, 109, 255}
\definecolor{cb-blue-sky}   {RGB}{109, 182, 255}
\definecolor{cb-blue-light} {RGB}{182, 219, 255}
\definecolor{cb-burgundy}   {RGB}{146,   0,   0}
\definecolor{cb-brown}      {RGB}{146,  73,   0}
\definecolor{cb-clay}       {RGB}{219, 209,   0}
\definecolor{cb-green-lime} {RGB}{ 36, 255,  36}
\definecolor{cb-yellow}     {RGB}{255, 255, 109}

\usepackage{stfloats}

\usepackage{inconsolata}
\usepackage{hyperref}
\title{The Classics at SemEval-2026 Task 3: Combining Transformer Models and LLM-Generated Annotations for Dimensional Aspect-Based Sentiment Analysis}

\usepackage{inconsolata}
\usepackage{hyperref}
\usepackage{fontawesome5}
\author{Rafif Alshawi \textsuperscript{\faCrow}\thanks{Equal contribution.}, 
        Amit Raj \textsuperscript{\faCrow}\footnotemark[1], 
        Aleksey Kudelya \textsuperscript{\faCrow}, 
        Alexander Shirnin\textsuperscript{\faCrow} \\
    \textsuperscript{\faCrow}HSE University \\
    \small{
    \textbf{Correspondence:} \href{mailto:ashirnin@hse.ru}{\texttt{ashirnin@hse.ru}}
}
}

\begin{document}
\maketitle
\begin{abstract}
This paper presents an approach to the SemEval-2026 Task 3: Dimensional Aspect-Based Sentiment Analysis. We investigate methods for moving beyond traditional categorical sentiment (e.g., positive or negative) to predict fine-grained, real-valued scores for sentiment "valence" (positivity) and "arousal" (intensity). We participate in two subtasks: predicting these scores for given aspects (Subtask 1) and extracting full sets of sentiment details, including aspects, categories, and opinions alongside their scores (Subtask 3). Our approach for the regression task involves a weighted ensemble of transformer-based encoder models. For the Russian language, we further enhance the input by using a large language model (LLM) to generate synthetic sentiment descriptions. For the extraction task, we fine-tune a decoder LLM to perform structured prediction, allowing the system to identify sentiment elements and estimate their numerical scores simultaneously.
\end{abstract}

\section{Introduction}

A major concern of traditional Sentiment Analysis models is their opacity and tendency to oversimplify complex human emotions into discrete and rigid categories. As these models are deployed in an increasing number of applications, the inability to capture nuanced user feedback has exposed systematic vulnerabilities in downstream decision-making~\cite{liu2012sentiment}. Consequently, moving beyond categorical labels to capture the true intensity and nature of opinions has been posed as a critical desideratum in the rapidly developing field of Natural Language Processing~\cite{9996141}. To this end, Dimensional Aspect-Based Sentiment Analysis (DimABSA) has emerged to map sentiment onto a continuous two-dimensional space of \textit{valence} and \textit{arousal}~\cite{russell1980circumplex}. Developing methods capable of such complex reasoning over text helps uncover deeper semantic relationships and improves the faithfulness of the models to the underlying human emotions.

SemEval-2026 Task 3~\cite{yu-etal-2026-semeval} focuses on this paradigm shift by providing a multilingual and multidomain benchmark for fine-grained sentiment evaluation. The shared task offers multiple subtasks that correspond to standard requirements in modern affective computing. These include predicting the continuous valence and arousal scores for a given aspect (Subtask 1) and extracting the complete set of sentiment rationales such as the aspect, category, opinion phrase, and their continuous scores (Subtask 3).

This paper presents methods for complex reasoning over text to address the challenges of fine-grained sentiment analysis. For Subtask 1, we employ a weighted ensemble of encoders. To mitigate vulnerabilities in the Russian track, we introduce a zero-shot augmentation technique where synthetic rationales generated by a LLM provide explicit descriptive context. For Subtask 3, we address the systematic faults of multi-step pipelines~\cite{jing-etal-2021-seeking} by proposing a unified generative framework. 

Our participation provided several key insights into how different models handle fine-grained sentiment data. We found that the generative LLM performed exceptionally well on the quadruplet prediction task, which was surprising given that encoders are usually considered more stable for regression. Additionally, our results in the Russian track show that using an LLM to generate descriptive sentiment context can significantly help smaller encoder models understand numerical ranges.

\section{Background}

The DimABSA shared task~\cite{yu-etal-2026-semeval} provides a benchmark for evaluating fine-grained sentiment across several domains, including hospitality, consumer electronics, and finance. The dataset for Track A consists of multilingual and multidomain samples \cite{lee2026dimabsabuildingmultilingualmultidomain}, with our experiments specifically addressing the English and Russian tracks. This corpus requires models to move beyond discrete sentiment classification to capture the intensity and positivity of opinions through continuous values. We focus our participation on two specific challenges: Dimensional Aspect Sentiment Regression (Subtask 1) and Dimensional Aspect Sentiment Quad Prediction (Subtask 3).

\paragraph{Task Formulation} The two subtasks represent different levels of complexity in sentiment analysis. In Subtask 1, the model is provided with a text and a specific aspect, and it must predict the associated valence and arousal. In Subtask 3, the model must perform an end-to-end extraction of all sentiment-bearing components. The formulations are illustrated as follows:

\begin{tcolorbox}[colback=gray!8,colframe=red!85,fontupper=\small,title=Subtask 1: DimASR Example]
\textbf{Input Text:} The battery life is amazing. \\ 
\textbf{Given Aspect:} battery life \\
\textbf{Output (VA):} \texttt{8.50\#7.20}
\end{tcolorbox}
\begin{tcolorbox}[colback=gray!8,colframe=red!83,fontupper=\small, title=Subtask 3: DimASQP Example]
\textbf{Input Text:} The battery life is amazing. \\ \textbf{Output (Quadruplet):} \\ (\textit{Aspect}: battery life, \textit{Category}: LAPTOP\#BATTERY, \textit{Opinion}:amazing, \textit{VA}: 8.50\#7.20)
\end{tcolorbox}







\paragraph{Performance Metrics} To account for the continuous nature of the predictions, the organizers utilize metrics that penalize the distance between predicted and ground-truth values. For Subtask 1, performance is measured using the Root Mean Square Error (RMSE) across both valence and arousal dimensions:

\begin{small}
\begin{equation}
\label{eq:rmse}
\mathrm{RMSE}_{VA} = \sqrt{\sum_{i=1}^N \frac{(V_p^{(i)} - V_g^{(i)})^2 + (A_p^{(i)} - A_g^{(i)})^2}{N}}
\end{equation}
\end{small}

where $V_p$ and $A_p$ are the predicted scores, $V_g$ and $A_g$ are the gold labels, and $N$ represents the number of test instances. 

For Subtask 3, the evaluation uses a Continuous F1-score ($\mathrm{cF1}$). This metric unifies categorical extraction with numerical regression. A predicted quadruplet is first checked for a categorical match (Aspect, Category, and Opinion). If the categories match, the prediction is assigned a score based on a Continuous True Positive ($\mathrm{cTP}$) calculation, which reduces a perfect score of 1 by the normalized Euclidean distance ($dist$) between the predicted and gold VA values:

\begin{equation}
\label{eq:ctp}
\mathrm{cTP} = 1 - \frac{\sqrt{(V_p - V_g)^2 + (A_p - A_g)^2}}{\sqrt{128}}
\end{equation}

The final $\mathrm{cF1}$ is the harmonic mean of continuous precision and recall, ensuring that models are rewarded for both identifying the correct sentiment elements and accurately estimating their emotional intensity. All predicted scores must be restricted to the range of [1, 9] and rounded to two decimal places.

\section{System Methodology}
\label{sec:methodology}

This section details the methodologies developed to address the complex reasoning tasks presented in DimABSA. To accommodate the different requirements of the two subtasks, we design two independent pipelines. First, we outline our discriminative ensemble approach for Dimensional Aspect Sentiment Regression. Next, we present a generative framework for Dimensional Aspect Sentiment Quad Prediction.

\subsection{Subtask 1: Dimensional Aspect Sentiment Regression}
\label{subsec:subtask1_method}
Subtask 1 requires the model to infer continuous valence and arousal scores given a specific aspect. We formulate this as a multi-target regression problem, where the overarching goal is to map the input text directly to a two-dimensional continuous sentiment space.

\paragraph{Encoder Ensemble} We fine-tune a set of transformer-based encoder models to predict the two continuous variables simultaneously. A linear regression head is placed on top of the final hidden state of the \texttt{[CLS]} token to output the valence and arousal values. To ensure the robustness of our predictions and mitigate the variance inherent in individual models, we aggregate the outputs using a weighted ensemble strategy. The weights assigned to each model are optimized empirically based on their individual Root Mean Square Error (RMSE) performance on the development set.

\paragraph{LLM-based Augmentation for Russian} The Russian language track presents additional complexities regarding the extraction of subtle sentiment nuances. To alleviate this and provide the underlying encoder models with stronger linguistic signals, we introduce a zero-shot data augmentation step relying on an instruct-tuned LLM. For each training instance, we prompt the decoder model with the original text and instruct it to produce a short preliminary rationale describing the sentiment in terms of pleasantness and intensity. Rather than acting as a numerical score, this generated text serves as explicit descriptive context. The synthetic rationale is then concatenated with the original input text before being processed by the encoder models. We provide the whole system prompt used for this generation step in Appendix~\ref{appendix:a1}.

\subsection{Subtask 3: Dimensional Aspect Sentiment Quad Prediction}
\label{subsec:subtask3_method}
Subtask 3 involves a more complex reasoning process, requiring the simultaneous extraction of categorical elements (aspect, category, opinion) and the prediction of continuous elements (valence and arousal). Rather than relying on a multi-step pipeline of distinct models, which often suffers from error propagation, we formulate this as a unified structured generation task.

\paragraph{Generative Extraction} We employ a decoder LLM, fine-tuning it specifically on task-specific, domain-based data. The model is trained to process the input text and directly generate a structured sequence containing the complete (Aspect, Category, Opinion, Valence-Arousal) quadruplets. 

By training the model to jointly estimate the valence and arousal scores alongside the extraction of the aspect and opinion terms, the decoder model learns the underlying relationship between the descriptive textual phrases and their corresponding emotional intensity. Consequently, the model maps the extracted opinion directly to the numerical valence-arousal space in a single forward pass. This single-model architecture successfully avoids the systematic faults and compounding errors often observed in standard systems where extraction and regression are treated as isolated steps. The full system prompt used for this task is provided in Appendix~\ref{appendix:a2}.

\section{Experiments}\label{sec:exps}

\paragraph{Overview} We design a series of experiments to evaluate the proposed methods for dimensional sentiment analysis. We investigate the performance of established transformer-based language models, selected based on their efficacy on standard natural language processing benchmarks. For the English track of Subtask 1, we utilize \textit{RoBERTa-Large}\footnote{\href{https://huggingface.co/FacebookAI/roberta-large}{\texttt{hf.co/FacebookAI/roberta-large}}}, \textit{RoBERTa-Base}\footnote{\href{https://huggingface.co/FacebookAI/roberta-base}{\texttt{hf.co/FacebookAI/roberta-base}}}~\cite{liu2019robertarobustlyoptimizedbert} and \textit{DeBERTaV3-Large}\footnote{\href{https://huggingface.co/microsoft/deberta-v3-large}{\texttt{hf.co/microsoft/deberta-v3-large}}}~\cite{he2023debertav} models from the transformers library~\cite{wolf-etal-2020-transformers}.

For the Russian track, we employ \textit{XLM-RoBERTa-Base}\footnote{\href{https://huggingface.co/FacebookAI/xlm-roberta-base}{\texttt{hf.co/FacebookAI/xlm-roberta-base}}} and \textit{XLM-RoBERTa-Large}\footnote{\href{https://huggingface.co/FacebookAI/xlm-roberta-large}{\texttt{hf.co/FacebookAI/xlm-roberta-large}}}~\cite{conneau-etal-2020-unsupervised} to accommodate the multilingual requirements of the dataset. First, we detail the fine-tuning procedures and hyperparameter selection for the encoder models. Second, we present the data augmentation strategy developed for the Russian track and outline the final ensemble configuration.

\subsection{Subtask 1: Dimensional Aspect Sentiment Regression}

\paragraph{Encoder Fine-Tuning} 
We fine-tune the encoder models to predict continuous valence and arousal scores. The models are optimized using the Adam optimizer~\cite{ADAM}. All model parameters are updated during training. Our preliminary experiments suggest that training beyond five epochs leads to overfitting on the training set. Therefore, we limit the maximum number of epochs to five in all subsequent experiments.

To ensure hyperparameter selection, we allocate 10\% of the provided training data as an internal validation set to monitor performance during the learning phase. The official development set is strictly reserved for final model evaluation and ensemble weighting. For the RoBERTa and XLM-RoBERTa architectures, we experimented with various learning rates and observed that a learning rate of \texttt{2e-5} yields the best results. For the DeBERTaV3-Large model, we utilize a learning rate of \texttt{5e-6}, adhering to the architectural recommendations of the original authors~\cite{he2023debertav}, which we also empirically verified to be optimal for our data.

\paragraph{Ensemble Strategy}
To formulate our final submission for the English track, we aggregate the predictions of the three models using a weighted average. To minimize the risk of overfitting the relatively small development dataset, we avoid an exhaustive combinatorial search for optimal ensemble weights. Instead, we assign weights of 0.35 to the RoBERTa models and 0.30 to DeBERTaV3-Large, reflecting their relative performance on the development set.

\paragraph{Data Augmentation}
Our preliminary experiments indicate that the proposed data augmentation technique does not yield significant improvements for the English track. Conversely, applying this method to the Russian dataset provides a measurable performance advantage. We utilize the \textit{Llama-3.2-3B-Instruct}\footnote{\href{https://huggingface.co/meta-llama/Llama-3.2-3B-Instruct}{\texttt{hf.co/meta-llama/Llama-3.2-3B-Instruct}}} to generate synthetic sentiment rationales for the training instances. We do not fine-tune it to avoid overfitting; it is used in a zero-shot setting. To ensure computational efficiency during this step, we generated the texts using the vLLM framework~\cite{kwon2023efficient}.

We experimented with various prompts to instruct the decoder model. Based on a manual review of the generated rationales, we select a prompt that ensures structural consistency and qualitative plausibility of the sentiment descriptions. The prompt design was further assisted by a proprietary LLM \textit{Claude-Sonnet-4.6}\footnote{\href{https://platform.claude.com/docs/en/about-claude/models}{\texttt{claude.com/docs/en/about-claude/models}}}.

\subsection{Subtask 3: Dimensional Aspect Sentiment Quad Prediction}

\paragraph{Decoder Fine-Tuning}
For the structured extraction of sentiment quadruplets, we investigate the efficacy of instruction-tuned generative models. Constrained by available computational resources, we prioritize a parameter-efficient fine-tuning strategy, namely Low-Rank Adaptation (LoRA) tuning~\cite{hu2021loralowrankadaptationlarge}. We utilize the \texttt{unsloth} library~\cite{unsloth} to accelerate the training process and reduce memory overhead. Our primary experiments employ~\textit{Llama-3.1-8B-bnb-4bit}\footnote{\href{https://huggingface.co/unsloth/Meta-Llama-3.1-8B-bnb-4bit}{\texttt{hf.co/Meta-Llama-3.1-8B-bnb-4bit}}} model. We configure the training procedure to 500 steps, utilizing a learning rate of \texttt{1e-4} and a batch size of 8. To establish a comparative baseline, we also evaluate the~\textit{DeepSeek-R1-Distill-Llama-8B-bnb-4bit}\footnote{\href{https://huggingface.co/unsloth/DeepSeek-R1-Distill-Llama-8B-bnb-4bit}{\texttt{hf.co/DeepSeek-R1-Distill-Llama-8B-bnb-4bit}}}~\cite{Guo_2025} under identical hyperparameters.

\paragraph{Ablation and Post-Competition Setup}
Given the continuous nature of the valence and arousal scores, we initially hypothesized that decoder models might lack the underlying capabilities required for precise numerical regression. To investigate this, we designed an ablation experiment featuring a hybrid pipeline. In this setup, a dedicated RoBERTa-Base encoder model, specifically trained for regression, is utilized to predict the continuous scores for the extracted text spans, thereby overriding the decoder model's numerical outputs. 

Furthermore, because an exhaustive hyperparameter search was not computationally feasible during the active phase of the shared task, we conducted a systematic post-competition study on the official test dataset. We investigate variations in the learning rate, the impact of alternative system prompts, and the efficacy of different base architectures to validate our initial architectural choices.

\section{Results and Discussion}\label{sec:results}

\subsection{Subtask 1 Results}


\begin{table}[!htbp]
    \centering
    \resizebox{\columnwidth}{!}{
    \begin{tabular}{lccc}
        \toprule
        \textbf{System} & \textbf{Language} & \textbf{Domain} & \textbf{Test RMSE} \\
        \midrule
        Our Final System & English & Restaurant & 1.23 \\
        Baseline (Kimi-K2 Thinking) & English & Restaurant & 2.14 \\ 
        Baseline (Qwen-3 14B) & English & Restaurant & 2.64 \\ 
        \midrule
        Our Final System & English & Laptop & 1.33 \\
        Baseline (Kimi-K2 Thinking) & English & Laptop & 2.18 \\ 
        Baseline (Qwen-3 14B) & English & Laptop & 2.80 \\ 
        \midrule
        Our Final System & Russian & Restaurant & 1.64 \\
        Baseline (Kimi-K2 Thinking) & Russian & Restaurant & 1.77 \\ 
        Baseline (Qwen-3 14B) & Russian & Restaurant & 2.15 \\ 
        \bottomrule
    \end{tabular}
    }
\caption{Performance of our final system on the Subtask 1 test set compared to the official organizer baselines. Lower RMSE values indicate better predictive accuracy.}
\label{tab:subtask1_test_results}
\end{table}

The performance of our individual regression models on the development set is detailed in Table \ref{tab:subtask1_results}. Upon evaluating the English track (restaurant domain), we observe RMSE scores of approximately 1.15 and 1.17 for the RoBERTa models, compared to 1.23 for DeBERTaV3-Large. The models behavior is similar in the laptop domain. We hypothesize that this performance disparity is closely tied to the textual nature of the dataset; the predominantly short sentence structures align favorably with RoBERTa's pre-training distribution, granting it an empirical advantage over DeBERTa in this specific context.

For the Russian track, empirical results demonstrate the effectiveness of our synthetic rationale generation. The inclusion of the augmented context reduces the RMSE of the XLM-RoBERTa models from the 1.43--1.45 range down to 1.40--1.41. Due to the limited size of the Russian development set, we prioritize robustness in our final submission. Relying on the four-model ensemble provides a stable consensus prediction that mitigates the variance of any single architecture.

\begin{table*}[htbp]
    \centering
    \footnotesize
	\begin{tabular}{ccccc}
		\toprule
		  \textbf{Language} & \textbf{Domain} &\textbf{Model Architecture} & \textbf{Data} & \textbf{Dev RMSE} \\
		\midrule
        English & Restaurant & RoBERTa-Base & Original & \textbf{1.15} \\
        English & Restaurant & RoBERTa-Large & Original & 1.17 \\
        English & Restaurant & DeBERTaV3-Large & Original & 1.23 \\
        \midrule
        English & Laptop & RoBERTa-Base & Original & \textbf{1.24} \\
        English & Laptop & RoBERTa-Large & Original & 1.27 \\
        English & Laptop & DeBERTaV3-Large & Original & 1.33 \\
        \midrule
        Russian & Restaurant & XLM-RoBERTa-Base & Original & 1.44 \\
        Russian & Restaurant & XLM-RoBERTa-Large & Original & 1.48 \\
        Russian & Restaurant & XLM-RoBERTa-Base & Augmented & \textbf{1.41} \\
        Russian & Restaurant & XLM-RoBERTa-Large & Augmented & 1.43 \\
		\bottomrule
	\end{tabular}
\caption{Performance of individual encoder models on the Subtask 1 development set. Lower RMSE values indicate better predictive accuracy. \textbf{Bold} numbers indicate the best performance for each task.}
\label{tab:subtask1_results}
\end{table*}

\subsection{Subtask 3 Results}

The outcomes of our extraction methodologies and post-competition analysis are presented in Table \ref{tab:subtask3_results}. Most notably, our ablation study reveals that the hybrid pipeline (yielding a cF1 of 0.3024) underperforms the end-to-end LLM strategy (0.3072). This contradicts our initial hypothesis that encoders are strictly superior for numerical regression, suggesting that the LLM successfully captures the relationship between the extracted textual terms and their emotional intensity.

Our post-competition study further validates the hyperparameter choices made during the active phase. We find that increasing the learning rate to \texttt{2e-4} decreases the overall cF1 score to 0.2917, indicating suboptimal convergence. Similarly, modifying the prompt structure to alternative formats severely degrades performance, resulting in a score of 0.2716, underscoring the high sensitivity of instruction-tuned LLMs to prompt phrasing. Finally, the DeepSeek-R1 distillation model exhibits inferior performance compared to the quantized Llama-3.1-8B baseline, achieving a score of 0.2960. Ultimately, the analysis confirms that our initial submission configuration remains a better solution for dimensional sentiment quad prediction.

\paragraph{Architectural Choices}
Reflecting on the differing architectures employed across the two subtasks, we address the possibility of applying a unified generative framework to the regression challenge in Subtask 1. During the system development phase, we prioritized Subtask 1 before addressing the Subtask 3. At that initial stage, we hypothesized that decoders might lack the underlying capabilities required for precise continuous numerical approximation. Consequently, we relied on established encoder ensembles equipped with regression heads to ensure robust predictions. However, our subsequent ablation experiments in Subtask 3 contradict this initial assumption. The empirical results demonstrate that instruction-tuned decoders successfully capture the relationship between textual terms and their emotional intensity without requiring a separate regression module. Given the strict timeline of the shared task and limited computational resources, we were unable to retrospectively adapt and fine-tune a unified generative framework for Subtask 1. Because the chosen LLM exhibits a capacity to predict continuous valence and arousal scores, extending this purely generative approach to standard dimensional regression represents a highly promising direction for future work.


\begin{table*}[htbp]
    \centering
	\footnotesize
	\begin{tabular}{lccc}
		\toprule
		  \textbf{Setup} & \textbf{Domain} & \textbf{Learning rate} & \textbf{Test cF1 Score} \\
		\midrule
        Llama-3.1-8B-Instruct & Laptop & \texttt{1e-4} & \textbf{0.3072} \\
        Llama + RoBERTa-Base & Laptop & \texttt{1e-4} and \texttt{2e-5} & 0.3024 \\
        Llama-3.1-8B-Instruct & Laptop & \texttt{2e-4} & 0.2917 \\
        Llama-3.1-8B-Instruct (Alternative Prompt) & Laptop & \texttt{1e-4} & 0.2716 \\
        DeepSeek-R1-Distill-Llama-8B & Laptop & \texttt{1e-4} & 0.2960 \\
        \midrule
        Baseline (Kimi-K2 Thinking) & Laptop & - & 0.2795 \\
        Baseline (Qwen-3 14B) & Laptop & - & 0.1529 \\
		\bottomrule
	\end{tabular}
\caption{Post-competition evaluation on the Subtask 3 official test set. The single LLM from our official submission achieves the highest performance across all tested configurations. Higher cF1 values indicate better predictive accuracy. \textbf{Bold} numbers indicate the best performance for the task.}
\label{tab:subtask3_results}
\end{table*}

\section{Conclusion}\label{sec:conclusion}

This paper presents our system submitted to SemEval-2026 Task 3 on Dimensional Aspect-Based Sentiment Analysis. Our solution demonstrates competitive efficacy, placing 10th and 11th in the English laptop and restaurant domains for Subtask 1, 16th in the Russian track, and securing 7th place out of 18 in Subtask 3. For the regression challenges of Subtask 1, we adopted a classic weighted ensemble of transformer-based encoders, which we further enhanced specifically in the Russian track by introducing zero-shot synthetic rationale generation. For the structured quad prediction in Subtask 3, we transitioned to a purely generative framework utilizing a LLM. The primary advantage of employing a LLM for this complex extraction is its capacity to capture the relationship between textual phrases and their corresponding emotional intensity.

We believe this methodology holds significant potential for advancing affective computing. Future work will investigate the scalability of synthetic rationale generation across diverse languages and aim to enhance the overall system robustness against severe domain-shift scenarios.

\section*{Limitations}
Despite these advances, our evaluation has been limited to relatively small models, leaving the performance of current proprietary LLMs unknown. Furthermore, it remains unclear whether these algorithms will scale effectively to larger datasets with a large number of domains and longer reviews. Another critical challenge is the sensitivity to hyperparameter choices and system prompt design, which demonstrate a significant impact on the predictions. Future work will explore these limitations, focusing on larger models while addressing challenges in hyperparameter selection.

\section*{Acknowledgments}
This work is an output of a research project (HSE-BR-2025-025) implemented as part of the Basic Research Program at HSE University. We acknowledge the computational resources of HSE University's HPC facilities.

\bibliography{custom}

\appendix

\section{Data augmentation visualization}

\begin{figure*}[!th]
    \centering
    \includegraphics[width=0.9\textwidth]{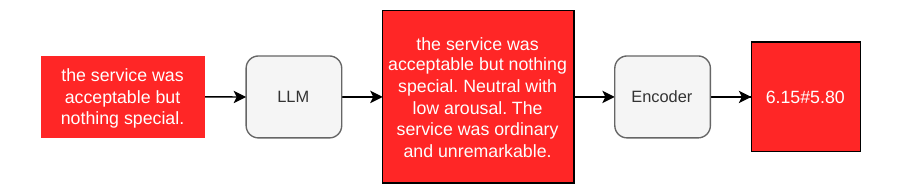}
    \caption{Visualization of the inference pipeline for Subtask 1 (Russian track). English text is used here for illustrative purposes.}
    \label{fig:appendix_attention}
\end{figure*}

We provide a schematic representation of the zero-shot data augmentation strategy employed for the Russian language track in Figure~\ref{fig:appendix_attention}. To facilitate readability and broader accessibility, the illustrative examples within the figure are presented in English, although the actual experiments were conducted on Russian instances. The pipeline depicts the generation of synthetic sentiment rationales via an instruct-tuned LLM and their subsequent concatenation with the original input text.

\section{LLM System Prompts}
\subsection{Prompt for Subtask 1}
\label{appendix:a1}

The following prompt is used to instruct the language model for generating explicit descriptive context:

\begin{tcolorbox}[
colback=gray!8,
colframe=red!85,
title=System Prompt, 
breakable,
fontupper=\small
]
You are an emotion analysis expert. Given a text and a specific aspect from that text, provide a brief emotional characterization that captures the sentiment toward that aspect.\\

Use the Valence-Arousal model:\\
- Valence: negative (1) $\leftrightarrow$ neutral (5) $\leftrightarrow$ positive (9)\\
- Arousal: low (1) $\leftrightarrow$ medium (5) $\leftrightarrow$ high (9)\\

Key emotion terms by quadrant:\\
High-Arousal Positive: delighted, excited, happy, thrilled, enthusiastic\\
Low-Arousal Positive: content, calm, relaxed, satisfied, pleasant\\
High-Arousal Negative: angry, tense, frustrated, annoyed, irritated\\
Low-Arousal Negative: depressed, bored, tired, disappointed, indifferent\\
Neutral/Mild: unremarkable, ordinary, neutral, forgettable, acceptable\\

Your task:\\
1. Analyze how the text describes or relates to the given aspect\\
2. First state the emotional classification using the terms above\\
3. Then provide one brief sentence describing the sentiment\\
4. Be precise - if sentiment is unclear or neutral, say so\\
5. Use English language only\\

Output format: Classification first, then description. Total 1-2 sentences, no additional explanation.\\

Examples:\\

Text: "their sake list was extensive, but we were looking for purple haze, which wasn't listed but made for us upon request!"\\
Aspect: "sake list"\\
Output: Positive and excited. The sake list is extensive and impressive.\\

Text: "the spicy tuna roll was unusually good and the rock shrimp tempura was awesome, great appetizer to share!"\\
Aspect: "spicy tuna roll"\\
Output: Highly positive and delighted. The roll is unusually good and exceeds expectations.\\

Text: "we love the pink pony."\\
Aspect: "pink pony"\\
Output: Positive and content. The restaurant is loved and valued.\\

Text: "this place has got to be the best japanese restaurant in the new york area."\\
Aspect: "place"\\
Output: Extremely positive and thrilled. The restaurant is considered the absolute best.\\

Text: "the service was acceptable but nothing special."\\
Aspect: "service"\\
Output: Neutral with low arousal. The service was ordinary and unremarkable.\\
\end{tcolorbox}

\subsection{Prompt for Subtask 3}
\label{appendix:a2}
The prompt below instructs the language model to predict valence and arousal scores and to extract aspect and opinion terms. \textbf{Bold} text corresponds to Markdown-style emphasis (i.e., **text**) present in the original prompt.

\begin{tcolorbox}[
colback=gray!8,
colframe=red!85,
title=System Prompt, 
breakable,
fontupper=\small
]
You are an expert Linguist specializing in Aspect-Based Sentiment Analysis (ABSA). Your task is to extract highly accurate (A, C, O, VA) quadruplets from the given text.\\

\textbf{Core Extraction Rules:}\\
1. \textbf{Aspect (A):} The specific feature or entity mentioned. Must match the input text casing exactly.\\
2. \textbf{Category (C):} Classify the aspect using the "ENTITY\#ATTRIBUTE" schema below. Use ONLY these labels. Must be UPPERCASE.\\
3. \textbf{Opinion (O):} The specific word/phrase used to express the sentiment. Match the input casing exactly.\\
4. \textbf{Valence-Arousal (VA):} \\
- \textbf{Valence:} 1.00 (Extremely Negative) to 9.00 (Extremely Positive). 5.00 is Neutral.\\
- \textbf{Arousal:} 1.00 (Calm/Sleepy) to 9.00 (Excited/Angry). 5.00 is Moderate.\\
- Format as "V.VV\#A.AA" (always 2 decimal places).\\

\textbf{Label Schema Constraints}:\\
- \textbf{Valid Entities}: {laptop\_entity}\\
- \textbf{Valid Attributes}: {laptop\_attribute}\\

\textbf{Step-by-Step Reasoning:}\\
Step 1: Identify all sentiment-bearing phrases and the aspects they refer to.\\
Step 2: Map each aspect to the most relevant ENTITY and ATTRIBUTE from the schema.\\
Step 3: Determine the numerical Valence (positivity) and Arousal (intensity) of the opinion.\\
Step 4: Format as a list of (A, C, O, VA) quadruplets.\\

---\\
\textbf{Examples:}\\

\textbf{Input:} The screen is incredibly bright and vibrant, but the price is a bit steep.\\
\textbf{Output:} (screen, DISPLAY\#QUALITY, incredibly bright and vibrant, 8.50\#7.20), (price, LAPTOP\#PRICE, a bit steep, 3.20\#5.50)\\

\textbf{Input:} The keyboard feels mushy and the battery drains too fast.\\
\textbf{Output:} (keyboard, KEYBOARD\#USABILITY, feels mushy, 3.00\#4.50), (battery, BATTERY\#OPERATION\_PERFORMANCE, drains too fast, 2.00\#6.50)\\

---\\
\textbf{Target Task:}\\
Input:\\
\texttt{\{input\_text\}}\\

Output:\\
\end{tcolorbox}

\end{document}